\documentclass[10pt, a4paper]{article}
\usepackage{lrec}
\usepackage{graphicx}
\usepackage{tabularx}
\usepackage{soul}
\usepackage{epstopdf}
\usepackage{xstring}
\usepackage{color}
\usepackage{dirtytalk}
\usepackage{times}
\usepackage{latexsym}
\usepackage{url}
\usepackage{dirtytalk}
\usepackage{xcolor}

\usepackage{natbib}
\setcitestyle{numbers,square}

\usepackage[utf8x]{inputenc}
\usepackage[T1]{fontenc}
\usepackage[thai,main=english]{babel}

\usepackage{fonts-tlwg}
\usepackage{setspace}
\usepackage{hyperref}

\renewcommand{\cite}{\citep}
\renewcommand{\cite}{\citet}

\title{Multiple Segmentations of Thai Sentences for Neural Machine Translation}
\name{Alberto Poncelas\textsuperscript{1}, Wichaya Pidchamook\textsuperscript{2}, Chao-Hong Liu\textsuperscript{3}, James Hadley\textsuperscript{4}, Andy Way\textsuperscript{1}}

\address{\textsuperscript{1}ADAPT Centre, School of Computing, Dublin City University, Ireland\\
    \textsuperscript{2}SALIS, Dublin City University, Ireland\\
    \textsuperscript{3}Iconic Translation Machines\\
    \textsuperscript{4}Trinity Centre for Literary and Cultural Translation, Trinity College Dublin, Ireland \\
        \{alberto.poncelas, andy.way\}@adaptcentre.ie\\
        ch.liu@acm.org, wichaya.pidchamook2@mail.dcu.ie, HADLEYJ@tcd.ie\\}

\abstract{
Thai is a low-resource language, so it is often the case that data is not available in sufficient quantities to train an Neural Machine Translation (NMT) model which perform to a high level of quality. In addition, the Thai script does not use white spaces to delimit the boundaries between words, which adds more complexity when building sequence to sequence models. In this work, we explore how to augment a set of English--Thai parallel data by replicating sentence-pairs with different word segmentation methods on Thai, as training data for NMT model training. Using different merge operations of Byte Pair Encoding, different segmentations of Thai sentences can be obtained. The experiments show that combining these datasets, performance is improved for NMT models trained with a dataset that has been split using a supervised splitting tool.
\\ \newline 
}

\begin{document}

\maketitleabstract

In Machine Translation (MT), low-resource languages are especially challenging as the amount of parallel data available to train models may not be enough to achieve high translation quality. One approach to mitigate this problem is to augment the training data with similar languages \citep{lakew2018neural} or artificially-generated text~\citep{sennrich2015improving,poncelas2018investigating}.

State-of-the-art Neural Machine Translation (NMT) models are based on the sequence-to-sequence framework (i.e. models are built using pairs of sequences as training data). Therefore, sentences are modeled as a sequence of tokens. As Thai is a \textit{scriptio continua} language (where whitespaces are not used to indicate the boundaries between words) sentences need to be segmented in the preprocessing step in order to be converted into sequences of tokens.

The segmentation of sentences is a well-known problem in Natural Language Processing (NLP), Thai is no exception. How Thai sentences should be split\footnote{In this work, we use interchangeably the term split or segmentation to refer to the process of dividing a word or a sentence into sub-units.} has been discussed at the linguistic level \citep{aroonmanakun2007thoughts} and different algorithms have been proposed \citep{haruechaiyasak2008comparative} including dictionary-based and machine learning-based approaches.

In this work, we use the problem of segmentation to our advantage. We propose using an unsupervised training method such as Byte Pair Encoding (BPE) to obtain different versions of the same sentence with different segmentations (by using different merge operations).

This leads to different parallel sets (each with different segmentation on the Thai side). These sets can be combined to train NMT models and achieve better translation performance than using one segmentation strategy alone.

%\section{Thai Language}
%\citet{aroonmanakun2007thoughts} discuss.......................

\section{Combination of Segmented Texts}

As the encoder-decoder framework deals with a sequence of tokens, a way to address the Thai language is to split the sentence into words (or tokens). We investigate three splitting strategies: (i) Character-based, using each character as token. This is the simplest approach as there is no need to identify the words; (ii) Supervised word-segmentation: using tokenizers which have been trained on supervised data such as the \say{Deepcut}\footnote{\url{https://github.com/rkcosmos/Deepcut}} library \citep{kittinaradorn2018thai} (trained on the BEST corpus \citep{kosawat2009best}); and (iii) Language-independent split: useing a language-independent algorithm for spliting such as BPE~\citep{sennrich2016neural}. This last method splits the sentence based on the statistics of sequences of letters. Therefore the split is not necessarily word-based. BPE is executed as follows: Initially it considers each character as an independent token; then iteratively the most frequent subsequent pair of tokens $x$ and $y$ are merged into a single token $xy$. The process is executed iteratively until a given number of merge operation are completed.

Once the sentences of a parallel corpus have been split, they can be used as training data for an NMT engine. Among the three approaches, the supervised word-segmentation strategy is the one that might allow NMT to perform the best, as it encodes the information from the manually-segmented sentences. However, by changing the merge operation parameter of BPE, the set can be split in different ways. Accordingly, in this work we want to explore whether augmenting the training set with the same sentences (but split with BPE using different merge operations) can be beneficial for training an NMT model. By doing this, we artificially increase the vocabulary on the target side. This causes the number of translation candidates of source-side words to be extended, which has been shown to have a positive impact in other approaches such as in statistical MT~\citep{poncelas2016extending,poncelas2017applying}. In addition, we want to explore whether this approach can be followed to augment a dataset split using Deepcut to improve the performance of NMT models.

Note that in this work we are building models in the English-to-Thai direction. The datasets with different splits are only on the target side whereas we keep the same number of merge operations on the source side. The main reason for this is that using several splits on the source side would also mean that NMT models would be evaluated with different versions (different splits) of the test set, which would obviously affect the results.

%we want to use a single test set for all the experiments. We propose as future work to use several split and the source side

%The reason for this limitation is that this way the test set is not conditioned by the different splits.

%\begin{itemize}
%    \item Do a concatenation of dataset with different language-independent split perform better than a language-dependent tokenizer?
%    
%    \item Do a concatenation of dataset with different language-independent language-dependent tokenizer to achieve better performance when used for training an NMT model?
%\end{itemize}

\section{Data and Model Configuration}

The models are built in the English to Thai direction using OpenNMT-py~\citep{opennmt}. We keep the default settings of OpenNMT-py: 2-layer LSTM with 500 hidden units and a maximum vocabulary size of 50000 words for each language.

We use the Asian Language Treebank (ALT) Parallel Corpus \citep{riza2016introduction} for training and evaluation. We split the corpus so we use 20K sentences for training and 106 sentences for development. We use Tatoeba~\cite{tiedemann2012parallel} for evaluating the models (173 sentences).

In order to evaluate the models we translate the test set and measure the quality of the translation using an automatic evaluation metric, which provides an estimation of how good the translation is by comparing it to a human-translated reference. As the evaluation is made in a Thai text (which does not contain {\em n}-grams, we use CHRF3~\citep{popovic2015chrf} which is a character-level metric, instead of BLEU~\citep{papineni2002bleu}, which is based on overlap of {\em n}-grams.

%We use the CHRF3~\citep{popovic2015chrf} metric because we want to use a character-level, instead of other popular {\em n}-gram level metric such as BLEU~\citep{papineni2002bleu}, because the evaluation is made in a Thai text (which does not contain {\em n}-grams).

The English side is tokenized and truecased (applying the proper case of the words) and we apply BPE with 89500 operations (the default explored in \citet{sennrich2016neural}).

%In the Thai side we explore different approaches for word segmentation. We build different dataset depending on the segmentation, but they can be classified as the following three:

For the Thai side we explore combinations of different approaches of sentence segmentation:

%. We build different dataset depending on the segmentation, but they can be classified as the following three:

%The Thai side is split following different approaches:
\begin{enumerate}
    \item Character-level: Split the sentences character-wise, so each character is a token of the sequence (\textit{character} dataset).
    
    %\item A Thai word tokenization library, Deepcut\footnote{\url{https://github.com/rkcosmos/Deepcut}} created by \citet{kittinaradorn2018thai} (\textit{Deepcut} dataset).% Note that this tool is language-dependent.
    \item Deepcut: Split the sentences using the Deepcut tool.
    
    \item Use BPE with different merge operations. In this work we explore with: 1000, 5000, 10000 and 20000 merge operations (\textit{BPE 1000}, \textit{BPE 5000}, \textit{BPE 10000} and \textit{BPE 20000} datasets, respectively). However, we propose as future work investigating the performance when using BPE with more merge operations.
    %A language-independent split using BPE: We explore different merge operations: 1000, 5000, 10000 and 20000.
\end{enumerate}

In total there are six different datasets. Note that they contain the same unique sentences. We train NMT models using these datasets either independently or in combination. The combination is carried out by appending different datasets. In particular we use \textit{character}, \textit{Deepcut} or \textit{BPE 1000} datasets and then accumulatively add \textit{BPE 1000}, \textit{BPE 5000}, \textit{BPE 10000} and \textit{BPE 20000} datasets.

\section{Experimental Results}

\begin{table}[!htbp]
\centering
\begin{center}
\begin{tabular}{ |p{3cm}|p{2cm}|}
\hline
	dataset	&	CHRF3	\\
\hline
Deepcut	&	47.90	\\
character	&	20.70	\\
BPE 1000	&	39.88	\\
BPE 5000	&	45.49	\\
BPE 10000	&	41.75	\\
BPE 20000	&	38.77	\\
\hline					
\end{tabular}
\caption{ 
Performance of the NMT model trained with data using different methods for word segmentation on the target-side.
}
\label{table:individual}
\end{center}
\end{table}

First, we evaluate the performance of the models when different merge operations are used on the Thai side. The performance of the model, evaluated using CHRF3 are presented in Table \ref{table:individual}. In the table, each row shows the results (the evaluation of the translation of the test set) of an NMT model built with the training set split following one of the thee chosen approaches. For example, the first row \textit{Deepcut} present the results when the model is built with the set after being split using the Deepcut tool, and in the row \textit{BPE 1000} if the dataset is split with BPE using 1,000 merge operations.

As expected, we see in the table that the best results are obtained when using the Deepcut tool (\textit{Deepcut} row), which splits Thai words grammatically.

Regarding the models in which BPE has been used, we see in Table \ref{table:individual} that the best performance is achieved when 5000 merge operations is used (\textit{BPE 5000} row). 

When too many merge operations are performed there is the risk of merging characters of different words into a single token, so the score decreases.

By contrast, if there are too few merge operations, the resulting text is closer to character-split which, as we can see in the \textit{character} row, leads to the worst results.

\subsection{Combination of Datasets with Unsupervised Split }

\begin{table}[!htbp]
\centering
\begin{center}
\begin{tabular}{ |p{3cm}|p{2cm}|}
\hline
	dataset	&	CHRF3	\\
\hline
\hline
character	&	20.70	\\
+ 1000	&	\bf38.23	\\
+ 5000	&	\bf45.63	\\
+ 10000	&	\bf49.93	\\
+ 20000	&	\bf52.17	\\
\hline
\hline
BPE 1000	&	39.88	\\
+ 5000	&	\bf49.54	\\
+ 10000	&	\bf52.33	\\
+ 20000	&	\bf53.45	\\
\hline
\end{tabular}
\caption{ 
Performance of the NMT model trained with data combining different unsupervised methods for word segmentation on the target-side. Numbers in bold indicate statistical significant improvement at p=0.01.
}
\label{table:combination}
\end{center}
\end{table}

The main focus of this paper is to study the impact on the performance when several training sets with different splits are gathered together. We show the results in Table \ref{table:combination}. Note that only the training set has been augmented, the development and test set remains the same for all experiments. In the table, each subtable presents the results when the sets are appended. For example, in the first subtable the row \textit{character} indicates the performance of the model trained with the data split by character (the same as in Table \ref{table:individual}). The following row, \textit{+1000}, shows the score of the model trained with character-wise split and that split with BPE using 1000 merge operations (so there are two instances of each source sentence). The last row of the subtable (\textit{+20000}) indicates the performance of the model trained with the five datasets concatenated (five instances of each source sentence).

In the table we observe that combining datasets with different segmentations is beneficial. In each subtable of Table \ref{table:combination} we see that the score goes up when we add more datasets with different splits. For example, the performance of the model using data split in a character-wise achieves a score of only 20.70 (first subtable). When we concatenate the same dataset but split using 1000 merge operations, the score increases to 38.23 (85\% improvement) and we see that the score increases as we add sentences with different splits. This effect is observed for all three subtables.

%is character-split beneficial.
The results also show that the lowest scores are achieved in those datasets in which a character-wise split fashion is performed. For example, the performance when using the dataset split with BPE using 1000 operations is 39.88 (\textit{BPE 1000} row in Table \ref{table:combination}), whereas when used in combination with character-split set (\textit{+1000} row in the first subtable of Table \ref{table:combination}) the score is 38.23. Another indication that character-wise splitting hurts performance is the comparison between the subtables of Table \ref{table:combination}. If we compare the rows \textit{+1000}, \textit{+5000}, \textit{+10000} and \textit{+20000} of each subtable, the lower scores are seen in the first subtable

Another research question we want to answer is whether using a combination of splits using BPE can outperform the model trained with a language-independent tool like Deepcut. We see that in the second subtable of Table \ref{table:combination} all of the the combination of BPE-split datasets (\textit{+5000}, \textit{+10000} and \textit{+20000} rows) exceed the 47.90 score achieved by using Deepcut alone.

\subsection{Augmenting the Dataset Split with Deepcut}

\begin{table}[!htbp]
\centering
\begin{center}
\begin{tabular}{ |p{3cm}|p{2cm}|}
\hline
	dataset	&	CHRF3	\\
\hline
\hline
Deepcut	&	47.90	\\
+ 1000	&	47.00	\\
+ 5000	&	\bf51.25	\\
+ 10000	&	\bf54.94	\\
+ 20000	&	\bf54.96	\\
\hline					
\end{tabular}
\caption{ 
Performance of combination of dataset with different split using \textit{Deepcut}. Numbers in bold indicate statistical significant improvement at p=0.01.
}
\label{table:combination_Deepcut}
\end{center}
\end{table}

As combining different splits can increase the performance of the model, does it also help when used in combination with Deepcut?

In Table \ref{table:combination_Deepcut} we present the results when the dataset split using Deepcut is augmented with datasets split with different merge operations using BPE. We see that initially, adding the set split with BPE using 1,000 merge operation (\textit{+1000} row) causes the performance to drop slightly. Nonetheless, the performance increases when more data are added (i.e. \textit{+5000}, \textit{+10000}, \textit{+20000} rows). Moreover, it even outperforms the model trained with the Deepcut-split dataset alone.

\subsection{Analysis}

%\begin{otherlanguage*}{thai}
%รถไฟมาแล้ว
%\end{otherlanguage*}
%\foreignlanguage{thai}{รถไฟมาแล้ว}
%\textthai{รถไฟมาแล้ว}

\begin{table*}[!htbp]
\centering
\begin{center}
\begin{tabular}{ |p{3.6cm}|p{5.2cm}|p{6.2cm}|}
\hline
	Dataset	&	\multicolumn{2}{|c|}{Output  }	\\
\hline
\hline
Source Sentence	&	\multicolumn{2}{|l|}{the train is here.  }		\\
Reference	&	\multicolumn{2}{|l|}{\textthai{รถไฟมาแล้ว}}		\\
\hline
Deepcut	&	\textthai{รถไฟนี้เป็นสายพันธุ์}	&	This train is a breed.		\\
\hline
BPE 20000	&	\textthai{รถไฟอยู่ที่นี่\textcolor{gray}{อยู่ที่นี่ด้วย}}	&	The train is located here.	\\
\hline
%Deepcut BPE 20000	&	\textthai{รถไฟนี้อยู่นี้}	&	This train is here.	\\
Deepcut + BPE 20000	&	\textthai{รถไฟนี้อยู่\textcolor{blue}{นี่}}	&	This train is located here.	\\	
\hline	
\hline	
Source Sentence	&	\multicolumn{2}{|l|}{I have a new bicycle.  }		\\
Reference	&	\multicolumn{2}{|l|}{\textthai{ฉันมีจักรยานใหม่}}		\\
\hline
Deepcut	&	\textthai{ผมได้ทำการทดลองใหม่}	&	I made a new experiment.		\\
\hline
BPE 20000	&	\textthai{ผมมีรถบรรทุก\textcolor{gray}{คน}ใหม่\textcolor{gray}{ๆ}}	&	I have a new truck.	\\
\hline
Deepcut + BPE 20000	&	\textthai{ผมมีรถจักรยานชุดใหม่\textcolor{gray}{ขึ้นมาแล้ว}}	&	I have a new set of bicycles.	\\
\hline
\hline
Source Sentence	&	\multicolumn{2}{|l|}{she does not have many friends in Kyoto.  }		\\
Reference	&	\multicolumn{2}{|l|}{\textthai{เธอไม่ค่อยมีเพื่อนมากนักที่เกียวโต}}		\\
\hline
Deepcut	&	\textthai{เธอยังไม่มีใครได้รับบาดเจ็บ}	&	She still doesn’t have anyone injured.		\\
\hline
%BPE 20000	&	\textthai{เธอไม่ได้มีเพื่อนหลายๆในเกียวโตเกียว}	&	She did not have many friends in Kyoto.	\\
BPE 20000	&	\textthai{เธอไม่ได้มีเพื่อนหลายๆในเกียวโต\textcolor{gray}{เกียว}}	&	She does not have many friends in Kyoto.	\\
\hline
%Deepcut BPE 20000	&	\textthai{เธอไม่มีเพื่อนร่วมหลายคนในเกียวโต}	&	She doesn’t have many friends together in Kyoto.	\\
Deepcut + BPE 20000	&	\textthai{เธอไม่มีเพื่อน\textcolor{gray}{ร่วม}หลายคนในเกียวโต}	&	She does not have many friends in Kyoto.	\\
\hline
\end{tabular}
\caption{ 
Examples of translations using different splits
}
\label{table:sentence_examples1}
\end{center}
\end{table*}

In Table \ref{table:sentence_examples1} we show some examples of the output. For each sentence we present the translations produced by the NMT model trained on \textit{Deepcut}, \textit{BPE 20000} and \textit{Deepcut BPE 20000}. We do not include the outputs of the NMT models trained on the \textit{character} dataset as all the translation consisted of the same, meaningless, sequence of characters. In the third column of the table we show the translation of the output of the NMT system after it has been postedited by removing characters (in gray) or replaced (in blue).

In the first sub-table we present how the sentence \say{the train is here} (meaning that the train is here because it has arrived) has been translated by different models. On the one hand, we see in the \textit{Deepcut} row that the model has not produced an accurate translation. On the other hand, the models trained with data containing sentences of different segmentations (i.e. \textit{BPE 20000} and \textit{Deepcut BPE 20000}) achieve a more accurate translation. Nevertheless, the output is not a perfect translation as it indicates where the train is located instead of expressing that it has arrived.

In the following subtables we observe a similar effect. The models trained with the \textit{Deepcut} dataset produced a translation that is either inaccurate or makes no sense. However, the models trained on \textit{BPE 20000} or \textit{Deepcut BPE 20000} produce a translation closer to the input after some post-editing (i.e. removing the characters in gray).

\section{Related Work}
\label{sec:related_work}

There are several studies aiming to address the problem of splitting words in Thai. One of the first approaches to segmenting is the \textit{longest-matching} method \citep{poowarawan1986dictionary}, consisting of identifying the longest sequence of characters that match a word in the dictionary. Another approach is \textit{maximal-matching} method \citep{sornlertlamvanich1993word}, which consists of generating all possible segmentations and retrieving those containing the smallest amount of words.

\citet{haruechaiyasak2009tlex} used conditional random fields to classify each character as either \textit{word-beginning} or \textit{intra-word}. \citet{nararatwong2018improving} proposed an improvement to this approach by adding information from POS tags.

The use of several segmentations has also been proposed by \citet{kudo2018subword}, in which he tries to integrate candidates from different segmentations. This technique has applications in a number of topics such as word-alignment \citep{xi2011word} or language modeling \citep{seng2009multiple,abate2010boosting}.

The use of multiple instances in the training data, where only one side is modified, has been used by \citet{poncelas2019combining}, who showed that using multiple instances of the same target sentences, with different source-sides (generated by different MT engines) leads to better results than using a single instance of each sentence.

\section{Conclusions and Future Work}
\label{sec:conclusions}

In this work we have explored whether the duplication of training sentences with different splits is useful to build NMT models with improved performance. The experiments show that the combination of different splits on the target side does improve NMT models involving a low-resource language such as Thai. 

The experiments also reveal that combining the same dataset using different merge operations of BPE not only improves the model trained on the dataset using the single configuration (regardless of the number of merge operations), but also the model trained on data that has been split using a tool trained on supervised data such as Deepcut.

In the future, we plan to conduct more fine-grained experiments to explore which configurations of BPE perform better. For example, would the combination of \textit{BPE 10000} and \textit{BPE 20000} (those with the highest number of operations explored) perform better than the model's original setup? And what would the results be if only \textit{BPE 1000} and \textit{BPE 20000} (those with the lowest and highest number of operation) are combined?

%we also want to explore whether these improvements can also be achieved when other languages are used. 

Furthermore, as all the experiments with different splits have been applied on the target side, we plan to investigate NMT models when Thai is on the source  side. Similarly, we will also investigate whether these improvements will be achieved using other languages.

Another variation we are interested in exploring is not to replicate all the sentences but to use data-selection algorithms to find a subset of sentences that may boost the performance of the models trained on the subset~\citep{poncelas2018data,poncelas2018adapt}.  

Finally, we would like to investigate the applicability of the method of employing several segmentations to other NLP tasks such as text classification \citep{apichai2019classifying,attaporn2019acquisition}.

\section{Acknowledgements}

The QuantiQual Project, generously funded by the Irish Research Council’s COALESCE scheme (COALESCE/2019/117).

This research has been supported by the ADAPT Centre for Digital Content Technology which is funded under the SFI Research Centres Programme (Grant 13/RC/2106).

%\section{Bibliographical References}\label{reference}

%\label{main:ref}
%\bibliographystyle{ksfh_nat}
%\bibliographystyle{plain}

%\bibliographystyle{lrec}
%\bibliographystyle{abbrvnat}
%\setcitestyle{numbers} 
\bibliography{lrec2020W-xample-kc}

%\printbibliography

\end{document}